\documentclass{article} % For LaTeX2e
\usepackage{iclr2025_conference,times}

\usepackage{amsmath,amsfonts,bm}

\def\eqref#1{equation~\ref{#1}}
\def\floor#1{\lfloor #1 \rfloor}
\def\1{\bm{1}}

\DeclareMathAlphabet{\mathsfit}{\encodingdefault}{\sfdefault}{m}{sl}
\SetMathAlphabet{\mathsfit}{bold}{\encodingdefault}{\sfdefault}{bx}{n}

\usepackage{todonotes}
\usepackage{url}
\usepackage{todonotes}
\usepackage[utf8]{inputenc} % allow utf-8 input
\usepackage[T1]{fontenc}    % use 8-bit T1 fonts
\usepackage[hidelinks,colorlinks]{hyperref}       % hyperlinks
\usepackage{url}            % simple URL typesetting
\usepackage{booktabs}       % professional-quality tables
\usepackage{amsfonts}       % blackboard math symbols
\usepackage{nicefrac}       % compact symbols for 1/2, etc.
\usepackage{microtype}      % microtypography
\usepackage{xcolor}         % colors
\usepackage{amsmath}
\usepackage[capitalise,noabbrev]{cleveref}
\usepackage{graphicx}
\usepackage{natbib}
\usepackage{inconsolata}
\usepackage{multirow}
\usepackage{tabularx}
\usepackage{enumitem}

\newcommand{\yes}{\textcolor{darkgreen}{\checkmark}}
\newcommand{\no}{\textcolor{darkred}{$\times$}}

\newcommand{\tabfn}[1][]{{\ifx&#1&\textsuperscript{$*$}\else\textsuperscript{\myfnsymbol{#1}}\fi}}
\newcommand{\myfnsymbol}[1]{\ensuremath{\ifcase#1 \or * \or \dagger \or \ddagger \or \mathsection \or \mathparagraph \else \@ctrerr \fi}}
\crefformat{footnote}{#2\footnotemark[#1]#3}
\crefrangeformat{footnote}{#3\footnotemark[#1]#4--#5\footnotemark[#2]#6}
\crefmultiformat{footnote}{#2\footnotemark[#1]#3}{\textsuperscript{,}#2\footnotemark[#1]#3}{\textsuperscript{,}#2\footnotemark[#1]#3}{\textsuperscript{,}#2\footnotemark[#1]#3}
\setlist{leftmargin=2em,topsep=0pt,partopsep=0pt,parsep=0pt,itemsep=3pt}
\definecolor{myblue}{rgb}{0,0.3,0.6}
\definecolor{darkgreen}{rgb}{0,0.3,0.1}
\definecolor{darkred}{rgb}{0.3,0,0}
\hypersetup{linkcolor=myblue,urlcolor=myblue,citecolor=myblue,anchorcolor=myblue}

\title{TeaserGen: Generating Teasers for Long\\Documentaries}
\author{\noindent
  \textbf{Weihan Xu}\textsuperscript{1} \quad  \textbf{Paul Pu Liang}\textsuperscript{2} \quad  \textbf{Haven Kim}\textsuperscript{3}   \quad  \textbf{Julian McAuley}\textsuperscript{3}\\[.5ex]
  \textbf{Taylor Berg-Kirkpatrick}\textsuperscript{3} \quad \textbf{Hao-Wen Dong}\textsuperscript{4}\\[1.5ex]
  \textsuperscript{1}Duke University \quad \textsuperscript{2}Massachusetts Institute of Technology\\
  \textsuperscript{3}University of California San Diego \quad \textsuperscript{4}University of Michigan\\[1.5ex]
  \tt\href{mailto:weihan.xu@duke.edu}{weihan.xu@duke.edu} \quad
  \href{mailto:ppliang@mit.edu}{ppliang@mit.edu} \quad
  \href{mailto:hak004@ucsd.edu}{hak004@ucsd.edu}\\
  \tt\href{mailto:jmcauley@ucsd.edu}{jmcauley@ucsd.edu} \quad
  \href{mailto:tberg@ucsd.edu}{tberg@ucsd.edu} \quad
  \href{mailto:hwdong@umich.edu}{hwdong@umich.edu}
}
\iclrfinalcopy % Uncomment for camera-ready version, but NOT for submission.
\begin{document}

\maketitle

\begin{abstract}
Teasers are an effective tool for promoting content in entertainment, commercial and educational fields. However, creating an effective teaser for long videos is challenging for it requires long-range multimodal modeling on the input videos, while necessitating maintaining audiovisual alignments, managing scene changes and preserving factual accuracy for the output teasers. Due to the lack of a publicly-available dataset, progress along this research direction has been hindered. In this work, we present
% requires long-range multimodal modeling capabilities
% Teasers are an important marketing tool, especially in the era of short videos. However, creating high-quality teasers often demands substantial manual editing. While teasers are widely used across different media, d
% Documentaries are commonly used as educational tools in academic and public settings. Teasers can serve as marketing tool to promote documentaries. However, generating teasers for long documentaries poses unique challenges in ensuring audiovisual correspondence, managing scene transitions and preserving factual accuracy at the same time.
% Further, the lack of a publicly available paired teaser and main contents makes generating teasers even more challenging. We present
\textit{DocumentaryNet}, a collection of 1,269 documentaries paired with their teasers, featuring multimodal data streams of video, speech, music, sound effects and narrations. With DocumentaryNet, we propose a new two-stage system for generating teasers from long documentaries.
The proposed \textit{TeaserGen} system first generates the teaser narration from the transcribed narration of the documentary using a pretrained large language model, and then selects the most relevant visual content to accompany the generated narration through language-vision models. For narration-video matching, we explore two approaches: a pretraining-based model using pretrained contrastive language-vision models and a deep sequential model that learns the mapping between the narrations and visuals. Our experimental results show that the pretraining-based approach is more effective at identifying relevant visual content than directly trained deep autoregressive models.
% that first summarizes the transcribed narration from the documentary and selects the most relevant visual content to accompany the summarized narration. For teaser narration generation,
% summarization,
% we adopt a pretrained large language model (LLM) to generate a teaser-like narration scripts with carefully designed prompts. For audiovisual matching, we first explore a pretraining-based audio-visual matching system \textit{TeaserGen-PT} using a pretrained model to find the most relevant visual content to the summarized narration through a thresholding mechanism. In addition we explore training a deep sequential model \textit{TeaserGen-LR} that learns the mapping between the narration and visual contents in documentaries. Our experimental results show that the proposed TeaserGen-PT model outperforms two existing models in a subjective test in terms of coherence, correspondence, engagingness and realness. 
% We also introduce two objective metrics \textit{Repetitiveness} and \textit{Scene Change Rate} that align human watching experience in terms of teaser. 
\end{abstract}

\section{Introduction}

Teasers are an effective tool for promoting video contents such as documentaries, movies, vlogs, commercials and educational videos. However, creating an effective teaser for long videos possesses unique challenges: first, it requires modeling and understanding long-range multimodal data streams of video, audio and narrations; second, it necessitates maintaining the text-visual correspondence between the teaser narrations and visuals; third, it needs managing smooth scene transitions beyond frame-by-frame audiovisual matching; finally, it entails preserving the factual accuracy in the generated teaser narrations and the accompanying visuals. These challenges together create an exciting yet underexplored ground toward long-range multimodal modeling.

Progress in teaser and trailer generation has been hindered due to the lack of a publicly-available dataset. Existing work on movie trailer generation \citep{movienet,mad,chi2024mmtrailmultimodaltrailervideo} relies on either private datasets or datasets without paired data, creating a barrier for the community to reproduce and follow up their research. In this paper, we present a new documentary dataset with 1,269 high-quality documentaries paired with their teasers. The proposed \textit{DocumentaryNet} dataset features various modalities such as video, speech, music, sound effects, narrations and tags. With the proposed dataset, we explore generating teasers for long documentaries.

In this work, we adopt a narration-centered approach for documentary teaser generation. Given a long documentary, we first generate the teaser narration from the transcribed narration, and then select the most relevant visual content from the documentary to accompany the generated teaser narration. We leverage a pretrained large language model (LLM) with specially-designed prompts to generate teaser narration that has attracting story-like narratives and a thought-provoking ending question. For narration-video matching, we first explore framing the task as a constrained optimization problem by applying a thresholding mechanism to maximize the total text-visual matching scores produced by a pretrained language-vision model, while keeping the total length of the selected video clips within a desired range. Further, we explore training a deep sequential model that directly learns the mapping between the narration and the visual content, and we propose various decoding strategies to maximize the narration-visual alignment while maintaining a cohesive output teaser. We compare our propose models with two baseline models \citep{lin2023univtg,clipit} using objective evaluation metrics and a subjective survey. The experimental results show that the text-visual matching score-based approach is more effective for narration-video matching compared to the directly trained sequential model.
Our contributions can be summarized as follows:
\begin{itemize}
\item We propose \textit{DocumentaryNet}, a publicly-available dataset consisting of 1,269 high-quality documentaries paired with their teasers.
% . The dataset includes modalities such as video, text, speech, music, and sound effects.
\item We propose \textit{TeaserGen}, a narration-centered teaser generation system that can effectively compress $>$30-min documentaries into $<$3-min teasers.
% two models that can effectively compress hours of video into multimodal teasers of less than 3 minutes. Both objective and subjective tests demonstrate the effectiveness of the proposed models.
% \item We introduce two new evaluation metrics, \textit{Repetitiveness} and \textit{Scene Change Rate}, which assess the quality of generated teasers from the perspective of the human viewing experience.
\end{itemize}

The DocumentaryNet dataset, all source code, pretrained models and hyperparameters will be made publicly available upon acceptance. Sample results can be found on our demo website.\footnote{\url{https://wx83.github.io/TeaserGen_Official/}\label{fn:demo}}

\begin{figure}
  \centering
  \includegraphics[width=\textwidth]{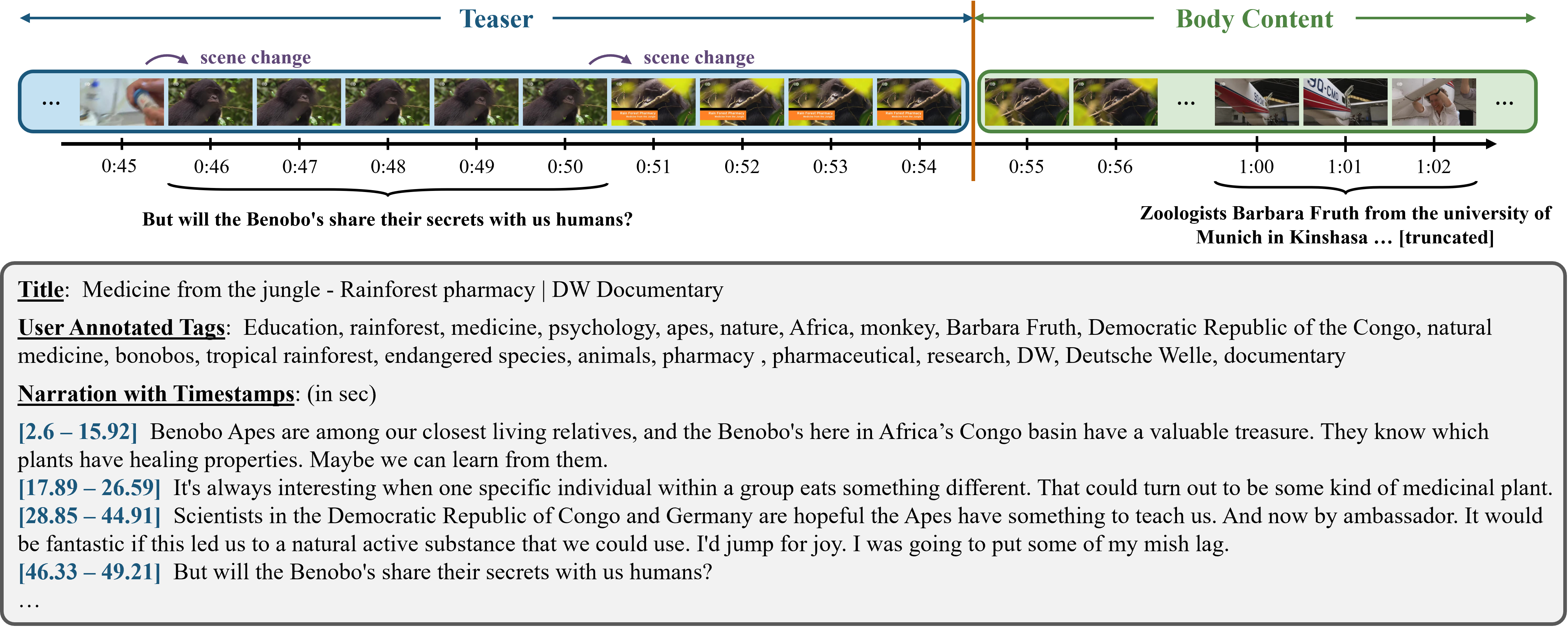}\vspace{-1ex}
  \caption{In this paper, we aim to generate teasers, including both the visual and narration, from a long documentary. This figure shows a sample from our \textit{DocumentaryNet} dataset, consisting of 1,269 high-quality documentaries paired with their teasers. } %Saliency curve Thresholding
  % \paul{not catchy enough, too much text and too few visuals}
  \label{fig:annotation}
\end{figure}

\section{Related Work}

\textbf{Teaser and trailer generation}\quad
% Some past
Prior work has studied
% considered
the task of teaser or trailer generation.  \cite{movietempo} proposed to identify and extract
% the most
semantically important
% significant
story units and segments of an action movie through movie tempo analysis. \cite{movietrailer2} learned to select key frames in a semi-supervised manner with a support vector machine. \cite{movietrailer3} learned to locate the shots that could be contributors to the trailer of an action movie with a support vector machine. However, those three methods do not account for temporal consistency throughout those frames.
\cite{argaw2024automatedmovietrailergeneration} proposed a unimodal sequence-to-sequence model that selects the key shots from movies without audio outputs.
% \tbf{while our work consider also the narration as inputs and outputs as narration is critical for documentaries}. 
CCANet \citep{wang2020ccanet} presented a ranking network that uses a co-attention mechanism between movies and trailers to guide the creation of training pairs, with moments strongly correlated with the trailers expected to receive higher scores than those that are less correlated. 
However, they do not attempt to generate audio narration or text captions to accompany the trailer with temporal correspondence, which is essential in documentary settings.
\cite{wang2024podreelshumanaicocreationvideo} proposed a human-AI co-creation system to assist video podcasters in crafting teasers. In this work, we propose a two-stage system that can output narration and visual contents with temporal correspondence. 

\begin{table}
    \scriptsize
    \centering
    \vspace{-3ex}
    \caption{Comparison of related methods in trailer generation and multimodal summarization }
    \vspace{1ex}
    \label{tab:method}
    \begin{tabular}{lcc@{~~~~}c@{~~~~}c@{~~~~}c@{~~~~}cc@{~~~~}c@{~~~~}cc}
        \toprule
        & &\multicolumn{5}{c}{Input modality} &\multicolumn{3}{c}{Output modality} &\multirow{2}{*}[-1ex]{\shortstack{Temporal multimodal \\information fusion}}\\
        \cmidrule(lr){3-7} \cmidrule(lr){8-10}
        &Type\tabfn & Frames &Video &Article &Narration &Query &Text &Video &Frames\\
        \midrule
        MMS (\citeyear{mms}) &E & \yes &\yes &\yes &\yes & \no &\yes & \no & \no &\no\\
        CLIP-IT (\citeyear{NEURIPS2021_7503cfac}) &E & \yes &\yes & \no &\no & \yes &\no &\yes & \yes &\no \\
        A2SuMM (\citeyear{he2023a2summ}) &E & \yes &\yes &\no &\yes &\no & \yes & \yes & \yes &\yes\\
        UniVTG (\citeyear{lin2023univtg}) &E  & \yes &\yes & \no &\no & \yes &\no &\yes & \yes &\no \\
        LfVS (\citeyear{argaw2024scalingvideosummarizationpretraining}) & E & \yes & \yes & \no & \yes  & \no & \yes & \yes & \yes & \yes \\
        TGT (\citeyear{argaw2024automatedmovietrailergeneration}) &E & \yes & \yes & \no & \no & \no & \no & \yes & \yes & \no \\
        \cmidrule(lr){1-11}
        MSMO (\citeyear{zhu-etal-2018-msmo}) &A & \yes & \no &\yes & \no & \no &\yes &\no & \yes &\no \\
        MM-AVS (\citeyear{mmavs}) & A &\yes & \yes & \yes &\yes & \no &\yes &\yes & \yes &\yes\\
        VideoXum (\citeyear{lin2023videoxum}) & A &\yes &\yes &\no & \no  &\no &\yes\tabfn[2] & \yes &\yes & \yes \\
        \cmidrule(lr){1-11}
        Ours & A &\yes & \yes & \no  & \yes & \yes &\yes &\yes & \yes &\yes\\
        \bottomrule
    \end{tabular}\\[.5ex]
    \tabfn \textbf{E}--Extractive; \textbf{A}--Abstractive \quad
    \tabfn[2] Achieved by dense video captioning \quad
    % \tabfn[3] Only included in pseudo summary generation
    \vspace{-1ex}
\end{table}

\textbf{Multimodal summarization}\quad
Multimodal summarization aims to create concise summaries by integrating information from multiple modalities. It can be categorized into several types based on how content is extracted and summarized.
The first category involves directly extracting text and visual content from the main material. For instance, \cite{argaw2024scalingvideosummarizationpretraining}  generates video summaries by leveraging both text and video inputs. They utilize large language models (LLMs) to extract key sentences from transcribed texts, which are then paired with time-aligned video segments to create pseudo-ground truth summaries. A2SuMM \citep{he2023a2summ} produces extractive summaries with a unified multimodal transformer-based model to predict key sentences and its time aligned video segments. Similarly, MMS \citep{mms} generates textual summaries from a set of documents, images, audios, and videos related to a specific topic. Another category of multimodal summarization focuses on obtaining textual information through dense video captioning. For example, \cite{lin2023videoxum} constructs the VideoXum dataset based on the video captioning dataset ActivityNet \citep{activitynet} and introduces an end-to-end cross-modal video summarization model, VTSUM-BLIP, to achieve video-to-video summarization, video-to-text summarization, and combined video-to-video and text summarization. However, they only input visual contents and generate text summary based on the encoded visual features. 
A third category involves converting abstractive summaries into extractive labels before identifying key sentences and frames. For instance, MM-AVS \citep{mmavs} introduces a Jump-Attention mechanism to align features between text and video by first converting abstractive summaries into extractive labels, then extracting key sentences and frames.
In our work, we generate teasers from long documentaries using both narration and visual content. Unlike traditional multimodal summarization, teaser generation requires handling longer input sequences, higher compression rates, and ensuring the narration remains engaging and story-driven.

% \paul{add section on cross-modal alignment work eg contrastive learning, dynamic time warping, see more in my survey paper}
\textbf{Cross-modal alignment}\quad 
Recent work in multimodal alignment includes discrete alignment and continuous alignment \citep{liang2023foundationstrendsmultimodalmachine}. Contrastive learning \citep{crossmodal1, crossmodal2, crossmodal3} is a common approach in discrete local alignment. Optimal transport-based approaches \citep{villani2009optimal} support global alignment between discrete elements across modalities. For continuous alignment, dynamic time warping \citep{dtw} can be used to segment and align multi-view time series data. In this work, we explore using a pretrained contrastive vision-language model as well as directly learning the cross-modal mapping between the narrations and visuals.

\section{Dataset}

We present \textit{DocumentaryNet}, a collection
% dataset consisting
of 1,269 high-quality documentaries from three reputable sources: DW Documentary, Public Broadcasting Service (PBS) and National Geographic. We download the videos and metadata from YouTube, where the metadata include
% capturing information such as the channel name, 
video title, duration, and user-annotated tags.
% In order to get
Each documentary in our dataset includes a short teaser at the beginning, which allows us to acquire teaser--documentary pairs through splitting the documentary into its teaser and main content, as shown in \cref{fig:annotation}.
We will refer to the main documentary section as the \textit{body content}.
% Since a teaser is always included at the beginning of each documentary in our dataset,
To find the boundary between each teaser and its body content, we recruit three annotators to mark the start of the body content in each documentary.
% to acquire paired documentary and their teasers. 
% \cref{fig:annotation} shows an example of our annotation.
% The teaser usually ends when the title of the documentary shows up and the narration starts to introduce the start of the main story.
The teasers and the body contents have an average length of 79 seconds and 31.3 minutes, respectively, resulting in a 4.15\% compression rate.
% The DocumentaryNet dataset will be made publicly available upon acceptance.
% The main documentary section has an average length of 31 minutes, while the teasers has an average length of 79 seconds.

% To acquire paired teaser and body content segments, we recruit three
% % independent
% annotators to mark the start of the main content in each documentary. 

To extract narrations from the documentaries, we first separate audio track to three tracks: dialogue, music and sound effects and then use a speech transcription model to transcribe narration from the dialogue track. We adopt a pretrained model \citep{audiosep} for sound separation. Since the main content audio is extensive, we split it into 60-second intervals before separating it into the three tracks. To ensure smooth transitions between segments, we apply a windowing function with a window size of 5\% based on the standard audio sampling rate of 44,100 Hz. To acquire frame--sentence pairs for training, we adopt Whisper \citep{timewhisper} to transcribe the narration from the dialogues track and estimate the timestamps for each sentence. We also include silence detection information in this dataset for future research. More details of the dataset can be found in \cref{datasetdetail}. 

\begin{table}
\scriptsize
\centering
\vspace{-3ex}
\caption{Comparison of relevant datasets for video summarization and movie trailer generation. Our dataset is the first publicly available dataset with music and sound effects modality. }
\vspace{1ex}
\label{tab:dataset}
\begin{tabular}{lc@{~~~~}c@{~~~~}c@{~~~~}crc@{~~~~}ccc}
    \toprule
    &\multicolumn{4}{c}{Modality} &\multirow{2}{*}[-1ex]{Samples} &\multicolumn{2}{c}{Mean duration (min)} &\multirow{2}{*}[-1ex]{\shortstack{Compression\\rate (\%)}} &\multirow{2}{*}[-1ex]{\shortstack{Publicly\\available}}\\
    \cmidrule(lr){2-5} \cmidrule(lr){7-8}
    &Visual &Text &Speech &Music\&SFX & &Main content &Teaser\\
    \midrule
    SumMe (\citeyear{SumMe}) &\yes &\no &\no &\no &25 &2.4 &0.73 &30.4 &\yes\\ 
    TVSum (\citeyear{TVSum}) &\yes &\no &\no &\no &50 &3.9 &1.17 &27.9 &\yes\\ 
    CNN (\citeyear{mmavs}) &\yes &\yes &\yes &\no &203 &2.1 &0.12 &5.7&\yes\\
    Daily Mail (\citeyear{mmavs}) &\yes &\yes &\yes &\no &1,970 &1.4  &0.05 &3.6 &\yes\\
    BLiSS (\citeyear{he2023a2summ}) &\yes &\yes &\no &\no &13,303 &- &0.17 &- &\yes\\
    MMSum (\citeyear{Qiu2023MMSumAD}) &\yes &\yes &\no &\no &5,100 &14.5 &0.13  &0.9 &\yes\\
    LfVS-T  (\citeyear{argaw2024scalingvideosummarizationpretraining}) &\yes &\yes &\no &\no &1,200 &12.2&- &- &\no\\ 
    VideoXum (\citeyear{lin2023videoxum}) &\yes &\yes &\no &\no &14,001 &- &- &13.6 &\yes\\ 
    \cmidrule(lr){1-10}   
    MovieNet (\citeyear{movienet}) &\yes &\yes &\yes &\yes &1,100 &116 &1.9&- &\no\\ 
    TGT (\citeyear{argaw2024automatedmovietrailergeneration}) &\yes &\yes &\yes &\yes &23,604 &- &- &- &\no\\
    \cmidrule(lr){1-10}
    \textbf{DocumentaryNet (ours)} &\yes &\yes &\yes &\yes &1,269 &31.3 &1.3 &4.15 &\yes\\
    % VideoXum~\citep{lin2023videoxum} & Extractive & Text, Video & Text, Video
    % \midrule
    % Ours & Abstractive & Text (story), Video & Text, Video, Music & Yes & No \\
    \bottomrule
\end{tabular}
% \tabfn These dataset are not publicly available.
% \quad \tabfn[2] In our task, some of the video frames cannot be matched to main contents due to shift
\end{table}

\section{Method}

Documentaries often rely on the narration to convey information, while the visual plays a supplementary role in strengthening the narrative. In this paper, we adopt a narration-centered approach by first generating the teaser narration from the transcribed narration of the documentary (\cref{sec:narration_generation}) and selecting the most relevant visual content to accompany the teaser narration (\cref{sec:thresholding,sec:learning}).

\subsection{Generating Teaser Narrations by Prompting a Large Language Model}
\label{sec:narration_generation}

% Documentaries in our dataset rely heavily on the narration to convey information, while the visuals supplement the narrative.
% Although extractive key sentences can fully preserve original information, simply extracting sentences from transcribed narration would be non-consistent and engaging, as shown in \cref{extractivesummary}. 
% Hence, we propose to adopt a two-stage approach to first generate a story-like narration using an LLM given long-form transcribed narration in main documentary and then select visual contents to accompany the narration.

We leverage a pretrained large language model to generate the teaser narration from the transcribed narration of the documentary. We then adopt a text-to-speech model~\citep{tts} to synthesize the generated script into audio narration. We note that extractive models often fail to construct a coherent narration due to the frequent inserted interviews in documentaries, as exemplified in \cref{narrationcomp}. 

% long narration!
To accommodate longer documentaries, we divide the transcribed narration of each main documentary into 10 segments, as the average transcript contains around 3,900 words. For each segment, we prompt GPT-4o \citep{gpt-4o} to generate a one-sentence summary, resulting in a total of ten single-sentence summaries per documentary.
% \footnote{\textit{Prompt}: Summarize the paragraph in one sentence.}
% make it story-like -- how this differs from text summarization
% To generate story-like narrations from the transcribed text of the main documentary, we design GPT prompts as shown in \cref{sec:gptprompt}.
Moreover, since documentary teasers often resemble a story and conclude with a thought-provoking question, we instruct GPT-4o \citep{gpt-4o} to rewrite the ten summarized sentences into
% act as a narrator and craft
a story-like narration based on the ten summarized sentences\footnote{We use the following prompt: ``Rewrite the paragraph into an engaging story opening in 10 sentences or less, keeping all names and avoiding being replaced by pronouns.''} and, further, propose an ending question to end the teaser narration.\footnote{We use the following prompt: ``Given the title and the provided summary, formulate one thought-provoking and concise question that relate directly to the summary.''}
% (see \cref{sec:gptprompt} for the prompt we used).
For TeaserGen-PT, as we query the model separately, we retain the names of characters rather than replacing them with pronouns in the story-like narration. We will examine the effectiveness of this approach in \cref{sec:ablation}.
% We include GPT prompts in \cref{gpt_prompts}

% % add a ending question
% Finally, we
% % prompt GPT-4o to produce a cohesive documentary-like narration from the ten single-sentence summary and
% prompt GPT-4o to generate a provoking ending question for the teaser.\footnote{\tba--prompt} \tbaa{missing a transition} To accompany the GPT-summarized narration track with corresponding visual content, we propose two new algorithms to select key visual contents from body contents.

\subsection{Selecting Accompanying Video Clips for the Teaser Narrations}
% \subsection{Interval-based Audiovisual Matching through Thresholding the Matching Score Curves}
\label{sec:thresholding}

% Teasers generally contain highlights in a long video to capture the attention from the audience. Therefore, we define a score function $f$ to measure the likelihood of one frame being considered as a highlight. 
% Let $V$ be a input video represented as a sequence of frames with sample every $t$ seconds, i.e V = $\{X_1, X_2, X_3, \dots, X_{n}\}$, we define $f(X)$ as an importance score 
To accompany the GPT-summarized narration track with corresponding visual content, we formulate this task as a sentence-by-sentence optimization problem that aims to find the most relevant and important video clips in the body contents to pair with each sentence in the generated narration.
% on maximizing the importance score. 
Let $\mathcal{V}$ be an input video represented as a sequence of frames sampled at a certain frame rate (1 frame per second in this work),
% with sample every $t$ seconds,
i.e  $ \mathcal{V} = \{\mathbf{x}_1, \mathbf{x}_2, \mathbf{x}_3, \dots, \mathbf{x}_{n}\}$ where $n$ is the total number of frames.
% $$X_{n}$ denotes a frame at time step $t_{n}$. 
Let $\mathcal{S}$ be the input narration represented as a sequence of sentences, i.e., $\mathcal{S} = \{s_1, s_2, \cdots, s_m\}$, where $m$ is the total number of sentences.
% , where $S_{k}$ represents the $k^{th}$ sentence from the generated narration script. 
We first synthesize each sentence $s_i$ into a waveform, and 
let $\tau_i$ be the duration of the synthesized speech of sentence $s_i$. Our goal is to find a number of video clips from the body content that can accompany sentence $s_i$ and have a total duration of $\tau_i$. In the following two sections, we will introduce two approaches we propose to tackle this problem.

\begin{figure}
  \centering
  \includegraphics[width=\linewidth]{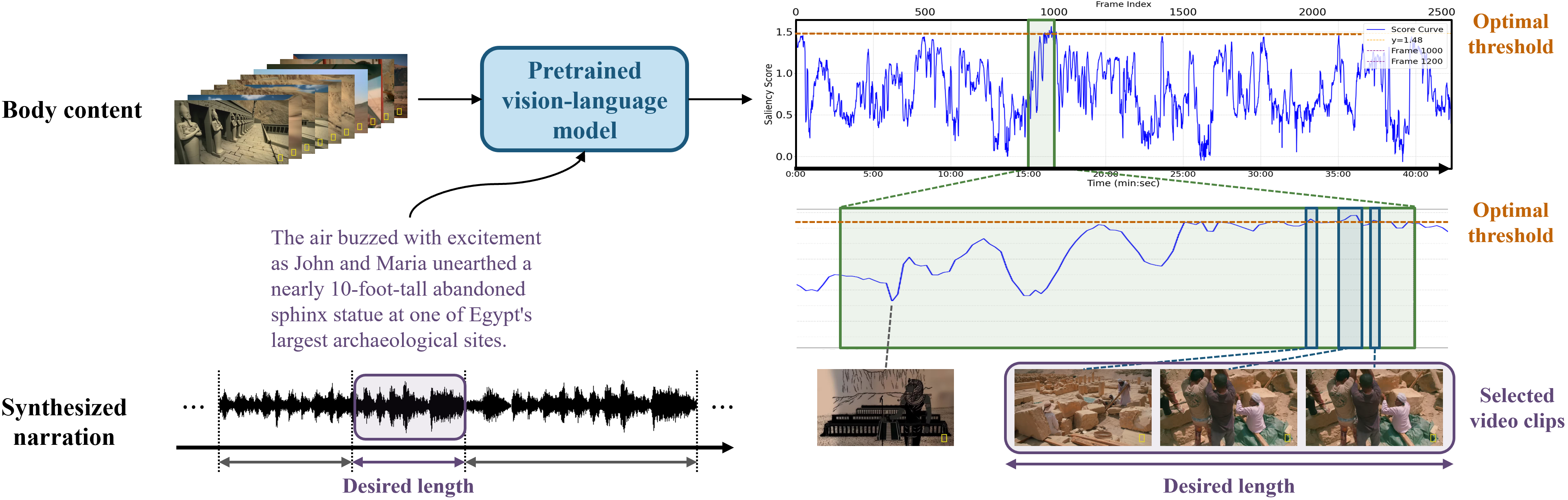}\vspace{-1ex}
  \caption{TeaserGen-PT selects video clips from the body content to accompany the generated narration by measuring the matching score between the input sentence and each video frame of the body content and applying a thresholding mechanism to find the optimal video clips. We apply binary search algorithm to find the optimal threshold. } %Saliency curve Thresholding
  \label{fig:method1}
\end{figure}

\subsubsection{Interval-based Approach using Pretrained Language-vision Models}
\label{sec:pretraining_based}

% \tbaa{probably remove this paragraph -- the math is too tricky to make it neat}
% % let $\tau = \{\tau_{1}, \tau_{2}, \cdots \tau_{m}\}$, where $\tau_{k}$ is the duration of synthesized speech corresponding to sentence $S_{k}$.
% We define $f(X_{i})$ as the importance score for frame $X_{i}$. 
% For each sentence, our objective is to select a number of 
% consecutive and important video frames that aligns the length of generated speech. Let $I_{k} = \{I_{k}^1, I_{k}^{2}, \cdots I_{k}^{M}\}$ represent the selected consecutive intervals of video frames for sentence $S_{k}$, where each $I_{k}^m$ is a set of consecutive video frames. The total length of all selected intervals should match $\tau_{k}$, i.e
% $\sum_{m=1}^{M} |I_{k}^m| \cdot \Delta t = \tau_k$ where $\Delta t$ is the time between frames and $|I_{k}^m|$ is the number of frames in the $m\text{-th}$ interval. The goal is to choose those intervals such that the sum of importance scores for all selected frames is maximized:
% \begin{equation}
%     \max_{I_k} \sum_{m=1}^{M} \sum_{X_{i} \in I_{k}^m} f(X_{i})
% \end{equation}

In the first approach, we frame this task as a constrained optimization problem, where we want to maximize the total text-visual correspondence between each sentence and the selected video clips while making the total length of the selected video clips be the length of the synthesized waveform of the sentence.
To achieve this, we use the VTGHL score (VTGHLS) produced by a pretrained video temporal grounding model \citep{lin2023univtg}, which measures the importance of the video clips (as defined by whether to be included as a highlight in video highlight detection) as well as the text-visual correspondence.
% initially tackle this problem by utilizing a pretrained video grounding model \citep{lin2023univtg} to define the importance, we refer to this importance score as VTGHLS.
% The importance is measured from both the visual-text alignment as well as whether the frame appears in foreground or background.
As shown in \cref{fig:method1}, we first generate a VTGHLS curve for each sentence throughout the main documentary, using a pretrained video temporal grounding model~\citep{lin2023univtg}. 
We then adopt binary search to find the threshold value that leads to the desired length. We will refer to this model as \textit{TeaserGen-PT}.

However, we notice this naive implementation leads to overly-fragmented videos with frequent scene changes and repeatedly selected video clips across sentences. To alleviate the overly-fragmented video issue, we only consider clips that are longer than 3 seconds to be selected in our algorithm. 
% when computing the total length in our binary search algorithm.
% However, we also notice that this naive search algorithm tends to select the same frames across different sentences, which is partly due to thSe nature of documentaries, as they focus on the same topic and subject throughout the video.
Further, to discourage the algorithm from selecting the same video clips across sentences, we only allow one second overlap with each of the video clips selected for the previous two sentences. We note that this issue is prominent in documentaries as a documentary focuses on the same topic and subject throughout the videos.
% discourage the algorithm from doing so, we only allow one second overlap with previous two selected intervals. 

\subsubsection{Learning-based Context-aware Approach using Deep Sequential Models}
\label{sec:learning}

\begin{figure*}
    \footnotesize
    \centering
    \includegraphics[width=.52\linewidth]{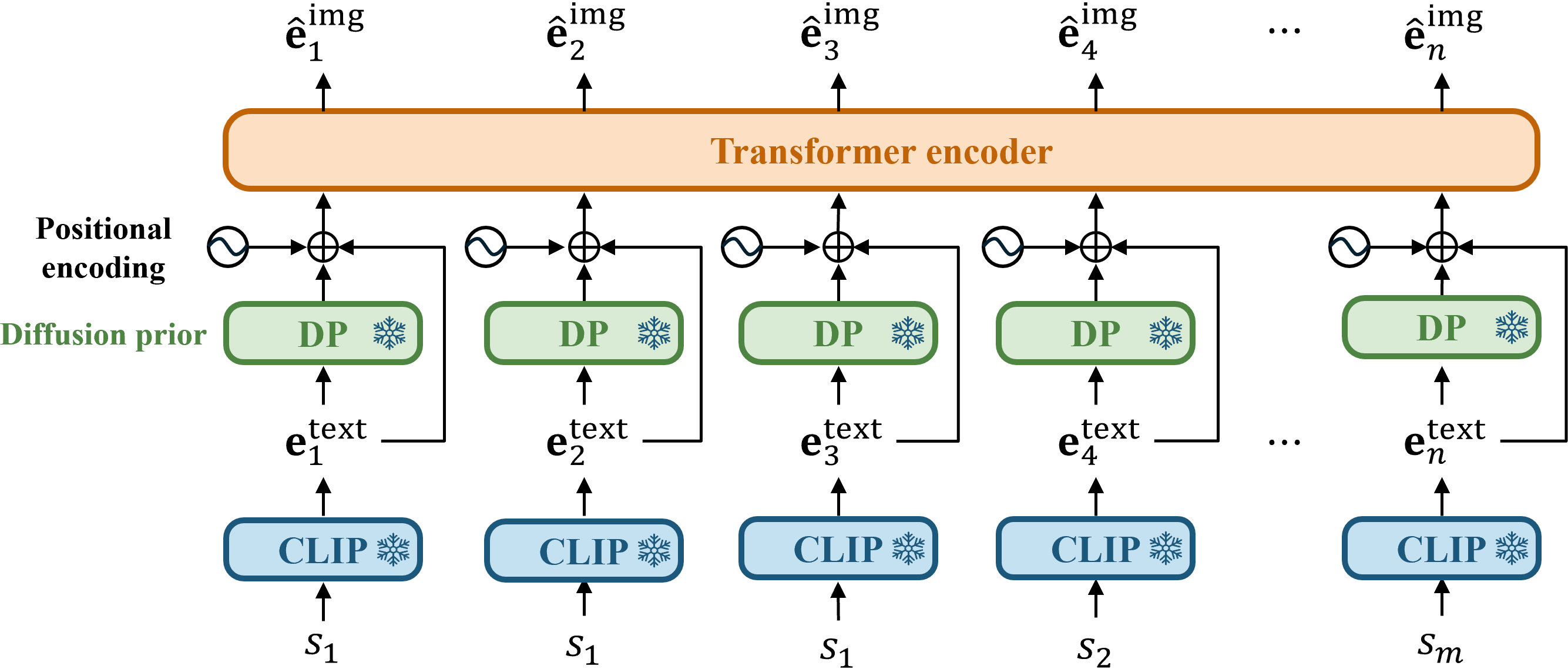}\hfill
    \includegraphics[width=.45\linewidth]{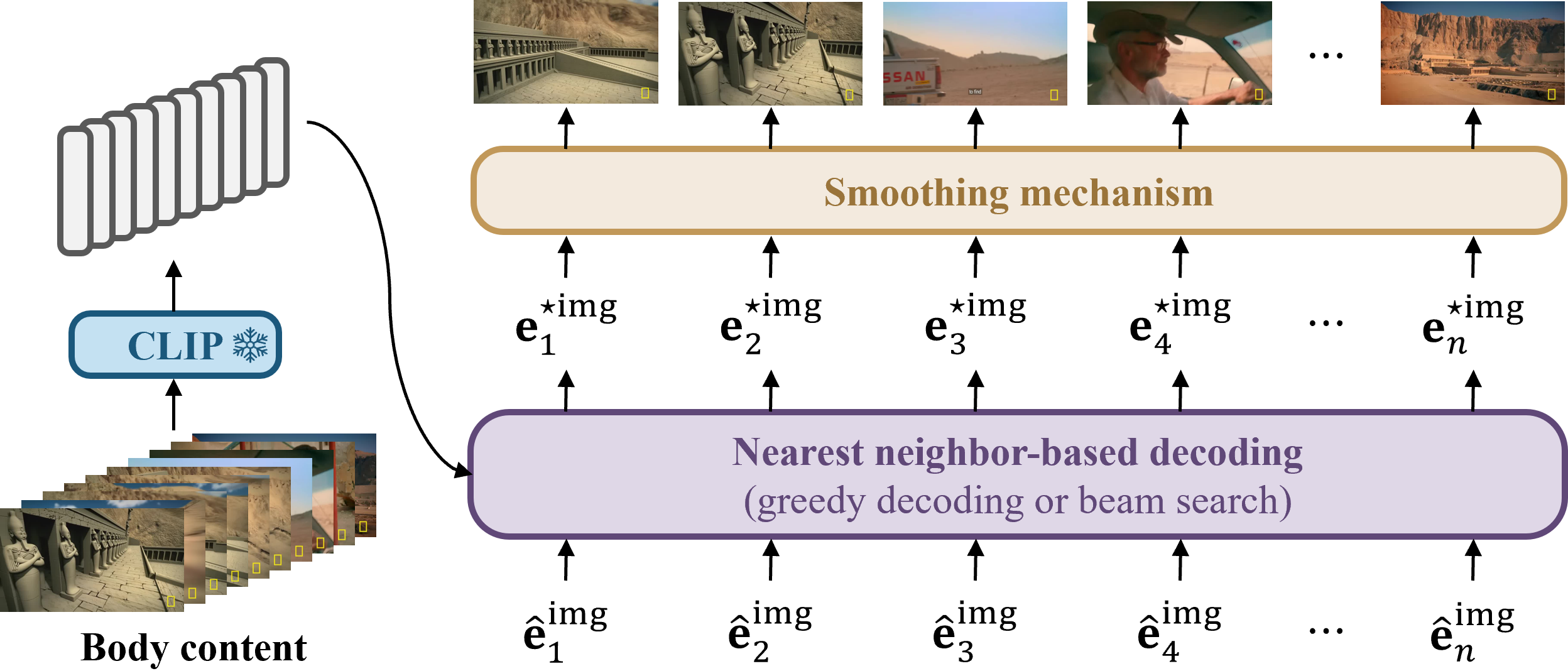}
    \caption{An overview of the proposed TeaserGen-LR model----(left) deep sequential model that learns the direct mapping between narration and visual content, and (right) the proposed decoding strategy using beam search and a smoothing mechanism}
    \label{fig:condition}
\end{figure*}

In the second approach, we adopt a train sequential model to learn the mapping between the narration and the visual content. We frame this as a sequence-to-sequence learning where the input is a sequence of sentence embeddings at each frame: $E^\mathrm{text} = (\mathbf{e}^\mathrm{text}_1, \mathbf{e}^\mathrm{text}_2, \cdots \mathbf{e}^\mathrm{text}_n)$, and the output is a sequence of the corresponding image embeddings: $E^\mathrm{img} = (\mathbf{e}^\mathrm{img}_1, \mathbf{e}^\mathrm{img}_2, \cdots \mathbf{e}^\mathrm{img}_n)$. In this work, we adopt a transformer model \citep{vaswani2023attentionneed} to learn the mapping $f: E^\mathrm{text} \rightarrow E^\mathrm{img}$. At inference time, we find the frame from the body content that has the closet embedding to the generated image embedding as the selected frame, as shown in \cref{fig:condition}. We will refer to this model as \textit{TeaserGen-LR}.

In practice, as the transformer operates at the frame level, multiple frames can share the same input sentence embeddings, and tend to generate similar output image embeddings within a sentence, which leads to overly-repeated scenes.
% since one sentence might correspond to multiple frames, those frames would have the same input embeddings.
% This might result in repetitiveness within one cut.
To alleviate this issue, we leverage a pretrained diffusion prior model \citep{ramesh2022dalle2,diffusionprior} that models the mapping from the CLIP-text embedding space to the CLIP-image embedding space. The diffusion prior model allows the model to generate more diverse output image embeddings,
% This diffusion prior samples image embeddings conditioned on the text, which are then added to the original text embeddings.
and, further, it helps reduce the modality gap between the CLIP image and text embedding spaces \citep{liang2022mindthegap}.

At inference time, we first explore decoding the output image embeddings by finding their nearest neighbors in the body content in the CLIP-image embedding space. In practice, however, we find that this naive approach leads to highly repetitive outputs. This is partly due to the nature of documentaries, which focus on the same topic and subject throughout the video, and thus frames at different parts of the documentaries may be semantically similar and have close CLIP-image embeddings.

To alleviate the overly-repetitive output issue across different sentences, we propose to add an additional regularization term to the decoding algorithm and adopt beam search to find the optimal sequence of frames. Let $\hat{E}^\mathrm{img} = (\mathbf{\hat{e}}^{\mathrm{img}}_1, \dots, \mathbf{\hat{e}}^{\mathrm{img}}_n)$ be the generated image embeddings from the model. Our goal is to find a sequence of frames in the body content that have the image embeddings $E^{\star\mathrm{img}} = (\mathbf{\hat{e}}^{\mathrm{img}}_1, \dots, \mathbf{\hat{e}}^{\mathrm{img}}_n)$ that maximize the following score function:
\begin{equation} \label{eq:score}
    \phi(\hat{E}^\mathrm{img}, E^{\star\mathrm{img}}) = \sum_{i = 1}^n d\left(\mathbf{\hat{e}}^{\mathrm{img}}_i, \mathbf{e}^{\star\mathrm{img}}_i\right) - \lambda\sum_{i, j\ \mathrm{in\ different\ sentences}} d\left(\mathbf{e}^{\star\mathrm{img}}_i, \mathbf{e}^{\star\mathrm{img}}_j\right)\,,
\end{equation}
where $d(\cdot, \cdot)$ is the distance metric (cosine similarity in this work) and $\lambda \in \mathbb{R}$ adjusts the strength of the regularization term. The regularization term in \cref{eq:score} encourage the algorithm to find diverse frames from the body content as the match to the raw output image embeddings. We apply beam search with a beam size $K_1$ to find the optimal decoded sequence. For computational efficiency, we consider only the top $K_2$ frames in the body content that are closest to the output embedding as candidates for selection. In this work, we use $K_1 = 5, K_2 = 10$. Finally, to prevent the model from selecting the same frames within one sentence, we also apply a smoothing mechanism that replaces $k$ repeatedly selected frames $\{\mathbf{x}_i, \dots, \mathbf{x}_i\}$ with nearby consecutive frames centered around $\mathbf{x}_i$:
$\{\mathbf{x}_{i - \floor{\frac{k}{2}}}, \dots, \mathbf{x}_{i + k - \floor{\frac{k}{2}} - 1}\}$.

\section{Experiments and Results}

\subsection{Implementation Details}\label{implementationdetail}

% We sample videos at a rate of one frame per second (1 fps) to represent the input videos.
We extract frames from the videos at a rate of one frame per second (1 fps).
For TeaserGen-PT, we use CLIP-ViT-B/32 \citep{clipl14} for extracting visual and sentence embeddings, chosen to match the embedding size of the pretrained video temporal grounding model. 
% \tbf{For Method 2,}
For TeaserGen-LR, we use CLIP-ViT-L/14 \citep{clipl14} for embedding extraction, aligning with the embedding size of the pretrained diffusion prior. The backbone transformer of the diffusion prior consists of 12 blocks with 12 attention heads. 
% \tbf{In Method 2,} 
For TeaserGen-LR, we utilize 3 transformer layers with a hidden dimension of 768, and we use the L2 distance between the ground truth image embeddings and the generated ones as the loss function. The batch size is set to 16, and we optimize using the Adam optimizer \citep{adam} with a learning rate of 1e-4. We train the proposed models for 15 epochs and select the best model on the validation set. During evaluation, we track the decoded frame numbers in the body content and use 512-dimensional CLIP embeddings for a fair comparison. We evaluate our model on a test set of 49 documentaries. All experiments are conducted on an NVIDIA RTX A6000 GPU.

\subsection{Baseline Models}

We consider four baseline models:\footnote{We intended to compare our model to \cite{argaw2024automatedmovietrailergeneration}. However, we cannot reproduce their results as it is trained on a private dataset, and their code and pretrained models are not available.}
\begin{itemize}
    \item
        \textbf{Random selection}: 
        This baseline model selects clips based on the duration of the generated speech. The model randomly selects $M$ clips from the main content, each corresponding to the duration of the synthesized speech for sentences $\{ s_1, \dots, s_m\}$.
        % Let $\tau = \{\tau_{1}, \tau_{2}, \cdots \tau_{k}\}$, where $\tau_{k}$ is the duration of synthesized speech corresponding to sentence $S_{k}$.
    \item \textbf{CLIP-NN}: This baseline CLIP nearest neighbor (CLIP-NN) model selects  key frames by finding frames in main documentary that are closet to the input sentences in CLIP embedding space. 
    \item
        \textbf{CLIP-IT} \citep{clipit}: 
        % \tbf{CLIP-IT \citep{clipit} is a language-based video summarization model. Since CLIP-IT cannot take input videos over 10 minutes, we divide the original video into 10 sub-clips and select key frames for each clip based on one sentence from GPT-generated scripts. We then concatenate all the selected frames in order.}
        This baseline model generates a teaser by selecting key frames based on any input query. Since CLIP-IT cannot process long video effectively, 
        we divide the original video into 10 sub-clips and pair each sub-clip with a sentence from the GPT-generated script. For each paired sub-clip and sentence, 
         % the duration of the synthesized speech of sentence
        suppose the duration of the synthesized speech of sentence is $\tau_i$, we extract top $\tau_i$ key frames that align with the assigned sentence. Finally, we concatenate all the selected frames in sequence to construct a teaser. 
    \item
        \textbf{UniVTG} \citep{lin2023univtg}: 
        This baseline model selects key shots with user-defined keywords or queries to construct a teaser. Since UniVTG cannot take input videos over 10 minutes, we adopt the same approach as in CLIP-IT to construct a teaser. 
\end{itemize}

% \paul{im worried someone will say these baselines are not enough or not recent}

% \textbf{Comparison to language-guided video summarization model}
 
\subsection{Objective Evaluation Metrics}

To evaluate the performance of our proposed methods, we adopt the following evaluation metrics:\footnote{Due to space concern, we provide details of the calculation in \cref{evaluationdetail}.\label{fn:evaluationdetail}}

\textbf{Retrieval-based metrics}\quad
Following \cite{smarttrailer, argaw2024automatedmovietrailergeneration}, we calculate the F1 score by comparing the frames selected in the generated teaser to those of the ground truth. While the F1 score provides us an estimate on the overlap between generated teaser and the ground truth, they should not be treated as the only standards as teaser generation is inherently a generative task.\cref{fn:evaluationdetail}
% After matching, we compare the frames selected by our model to the ground truth frames to compute these metrics.

\textbf{Repetitiveness}\quad
To measure the repetitiveness of generated teaser, we propose a new metric: 
\begin{equation}
    \mathrm{REP} = 1 - \frac{\text{Number of unique frames}}{\text{Number of frames}}
\end{equation}
The repetitiveness score computed on the ground truth is 7.86\%, suggesting that there are some repeated frames in the ground truth teasers.\cref{fn:evaluationdetail}
% We also provide an estimate of the ground truth Repetitiveness score as 7.86.
% \tbaa{so what are the values!?}
% , with
% Details of the calculation can be found in \cref{evaluationdetail}.
    
\textbf{Scene change rate (SCR)}\quad
To measure the degree of fragmentation, we propose the SCR metric:
\begin{equation}
    \mathrm{SCR} = 1 - \frac{\text{Number of consecutive frames within the same scene}}{{\text{Number of frames}}}
\end{equation}
SCR measures the frequency of scene changes of the teasers. The estimated SCR of the ground truth teasers is 27.6\% by manual inspections on 10 video samples.\cref{fn:evaluationdetail}
% We treat consecutive frames and repeated clips as a single clip. Individual frames are also considered as single clips.
% ground truth Scene Change Rate

\textbf{CLIPScore}\quad
To evaluate the alignment between the narration and the visual content, we compute the CLIPScore \citep{CLIPS}. Let $\mathbf{e}^{\mathrm{img}}_k$ and $\mathbf{e}^{\mathrm{text}}_k$ be the text and image embeddings at frame $k$, and $\mathrm{cos\_sim}(\cdot)$ be the cosine similarity. CLIPScore is defined as
\begin{equation}
    \mathrm{CLIPScore} = 2.5 \cdot \frac{1}{n} \cdot \sum_k^n \mathrm{max}\left(\mathrm{cos\_sim}\left(\mathbf{e}^{\mathrm{img}}_k, \mathbf{e}^{\mathrm{text}}_k\right), 0\right)\,,\vspace{-1ex}
\end{equation}
% both CLIPScore (CS) and CLIPScoreS (CSS) \citep{CLIPS} where
% \begin{equation}
%     \mathrm{CS} = \sum_i \mathrm{cos\_sim}(\mathbf{e}^{\mathrm{img}}_k, \mathbf{e}^{\mathrm{text}}_k)\,, \qquad
%     \mathrm{CSS} = 2.5 \sum_i \mathrm{max}(\mathrm{cos\_sim}(\mathbf{e}^{\mathrm{img}}_k, \mathbf{e}^{\mathrm{text}}_k), 0)\,.
% \end{equation}
% \tbf{very odd choice of symbols -- why c and v??} 

\textbf{VTGHLS}\quad
% \tbaa{this sentence needs rewriting -- can't follow the logic}
Proposed by \cite{lin2023univtg}, VTGHLS measures the importance of the frame (as defined by whether to be included as a highlight in video highlight detection) in addition to the text-visual correspondence. We estimate a VTGHLS of 0.64 for the ground truth teasers.\cref{fn:evaluationdetail}

\subsection{Objective Evaluation Results}
% compare with baseline

As shown in \cref{tab:gpt_objective}, the proposed TeaserGen-PT model that uses the title as the query outperforms both baseline models
% by a large margin,
as indicated by the higher F1 score and a closer scene change rate to that of the ground truth.
% that is more closely aligned with the ground truth.
Similarly, TeaserGen-LR, when using beam search, achieves a higher F1 score than than the baseline models and a closer scene change rate to that of the ground truth.
% surpasses the baselines with a higher F1 score and a scene change rate that also better reflects the ground truth.
In contrast, the two baseline models, i.e., UniVTG \citep{lin2023univtg} and CLIP-it \citep{clipit}, both lead to a scene change rate that is three to four times greater than that of the ground truth. Further, while the CLIP-NN baseline achieves the highest CLIPScore, it results in high repetitiveness, partly because the nearest neighbor search can easily lead to repeatedly selected video clips.
% CLIPScore is the optimization goal. The CLIP model also has high repetitiveness.
Moreover, TeaserGen-LR equipped with the diffusion prior model achieves us the second highest CLIPScore, demonstrating the effectiveness of diffusion prior in bridging the gap between the CLIP image and text embedding spaces. We also observe that transformer-based models (TeaserGen-LR) tends to result in a higher scene change rate than the pretraining-based models (TeaserGen-PT).

For TeaserGen-PT models, while it achieves a higher CLIPScore when using narration as the query, it achieves a higher F1 score when using title as the query.
% though using narration as the query results in a higher CLIPScore, TeaserGen-PT, when using the title as the query, achieves a higher F1 score.
This is possibly because a documentary often revolves around a specific topic or character, which is usually captured by the title,
% the title better encapsulates,
whereas narrations provide more detailed but denser descriptions.
For TeaserGen-LR models, in terms of F1 score, TeaserGen-LR, with beam search and diffusion prior, outperforms other transformer-based models. 
% We find that applying smoothing to encourage interval level correspondence to decrease the repetitiveness.
% Decoding with beam search method to discourage the selection of clips from different scenes can effectively decrease the repetitiveness. 
We find that decoding with the beam search method proposed in \cref{sec:learning} can effectively reduce repetitiveness by discouraging the selection of clips from different scenes.

\begin{table}
\scriptsize
\centering
\vspace{-3ex}
\caption{Objective evaluation results with LLM-generated narrations.  }
% \paul{way too many details in this one table}
\vspace{1ex}
\label{tab:gpt_objective}
\begin{tabular}{lllcccccccc}
\toprule
Model &Query &Decoding &DP &F1 (\%)$\uparrow$ &REP (\%) &SCR (\%) &CLIPScore &VTGHLS\\
\midrule
\textbf{Baseline models}\\
Random &Random &- &- &1.67  &4.05 &7.81 &0.56 &0.75\\
CLIP-NN &Narration &Greedy &\no &0.11 &92.73 &8.29 &0.69 &0.79\\
UniVTG (\citeyear{lin2023univtg}) &Title &Rank &- &1.82 &0 &89.68 &0.58 &1.01\\
CLIP-it (\citeyear{clipit}) &Narration &Rank &\no &1.24 &0 &99.39 &0.56 &0.61\\
\cmidrule(lr){1-9}
\textbf{Pretraining-based models}\\
TeaserGen-PT &Title &Thresholding &- &1.85 &0 &13.16 &0.56 &1.02\\
TeaserGen-PT &Narration &Thresholding &- &1.07 &21.38 &22.58 &0.58 &1.45\\
TeaserGen-PT-CLIP &Narration &Threshold &\no &1.31 &27.23 &24.10 &0.58 &0.74\\
\cmidrule(lr){1-9}
\textbf{Learning-based models}\\
% TeaserGen-LR &Narration &Greedy-Unsmoothed &\no &1.93 &0.31 &0.53 &81.32 &27.38 &0.24 &0.74 &0.59\\
TeaserGen-LR &Narration &Greedy &\no &1.56 &31.97 &27.18 &0.58 &0.74\\
TeaserGen-LR &Narration &Greedy &\yes &1.38 &26.83 &35.48 &0.62 &0.78\\
TeaserGen-LR &Narration &Beam search &\no &\textbf{1.88} & 24.16 &41.97 &0.58 &0.74\\
TeaserGen-LR &Narration &Beam Search &\yes & \textbf{1.88 } & 19.39 &46.56 &0.63 &0.77\\
% TeaserGen-LR &Narration & Threshold &\yes & 6.1896 &0.8917 &1.5589 &15.16 &23.10 &0.2531 &0.7631 &0.6329\\
\cmidrule(lr){1-9}
Ground truth &- &- &- &100 &$>$7.86 &27.6 &0.58 &0.64\\
\bottomrule
\end{tabular}
% \\[1ex]
% \tabfn \tbaa{did we want to say something here??} \quad 
% \tabfn[2] We manually inspect 10 videos to estimate the scene change rate for the ground truth teasers. 
\end{table}

\begin{table}
\footnotesize
\centering
\vspace{-3ex}
\caption{Subjective evaluation result with LLM-generated narrations.}
\vspace{1ex}
\label{tab:gpt_subjective}
\begin{tabular}{lllc@{~~~~}c@{~~~~}c@{~~~~}c}
\toprule
Model &Query &Decoding &Coherence$\uparrow$ &Alignment$\uparrow$ &Engagingness$\uparrow$ &Realness$\uparrow$\\
\midrule
UniVTG {\scriptsize(\citeyear{lin2023univtg})} &Title &Rank &2.61 $\pm$ 0.50 &2.62 $\pm$ 0.47 &2.67 $\pm$ 0.57 &2.66 $\pm$ 0.54\\
CLIP-it {\scriptsize(\citeyear{clipit})} &Narration &Rank &2.61 $\pm$ 0.46 &2.67 $\pm$ 0.44 &2.57 $\pm$ 0.46 &2.51 $\pm$ 0.46\\
% CLIP & Original & Narration & \no & 2.6 & 0.1 & 1.5 & \textbf{TODO} & 10 &  4.3 \\
\cmidrule(lr){1-7}
TeaserGen-PT &Title &Threshold &\textbf{3.14 $\pm$ 0.50} &2.84 $\pm$ 0.57 &\textbf{2.81 $\pm$ 0.49} &\textbf{2.94 $\pm$ 0.50}\\
TeaserGen-LR &Narration &Greedy &2.90 $\pm$ 0.45 &\textbf{2.88 $\pm$ 0.48} &2.71 $\pm$ 0.42 &2.71 $\pm$ 0.44\\
% TeaserGen-LR &Narration &Beam search & 2.75 $\pm$ 0.48 & 2.52 $\pm$ 0.48 & 2.71 $\pm $ 0.45 & 2.60 $\pm$ 0.47\\
TeaserGen-LR &Narration &Beam search &2.84 $\pm $ 0.46 &2.69 $\pm$ 0.51 & 2.71 $\pm$ 0.42 & 2.64 $\pm$ 0.41\\
\bottomrule
\end{tabular}
\end{table}

\subsection{Subjective Evaluation}
\label{sec:subjective}

To further measure the quality of our generated scripts and teasers, we conduct a subjective test. We randomly select 10 documentaries from the test dataset and split them into 2 versions, with each version containing demos for 5 documentaries. We then assess the coherence, video-narration alignment, engagingness, and realness (see \cref{sec:evaluation_questions} for the questions we ask in the survey) of the generated teasers on a Likert scale from 1 to 5. Additionally, we ask participants to compare the effectiveness of two approaches: summarizing each chunk of content directly versus using our carefully designed GPT prompts, also on a Likert scale from 1 to 5. This comparison focuses on how both methods perform in terms of consistency, informativeness, and engagingness. We recruit 20 participants for the subjective test, with 11 working on version A and 9 working on version B. Among all participants, 14 out of 20 have video editing experience.
% We include the human evaluation question in \cref{humaneval}

We report the subjective evaluation results in \cref{tab:gpt_subjective}. 
% along with 95\% confidence intervals.
Human evaluations further confirm that TeaserGen-PT surpasses baseline models with higher scores in terms of coherence, alignment, engagingness, and realness. The low coherence of baseline model in human study further proves that the baseline model is overly-fragmented. While TeaserGen-LR with beam search decoding has a higher F1 score and CLIPScore compared to TeaserGen-PT using the title as the query in objective metrics, subjective evaluations reveal that it performs lower in terms of coherence, alignment, engagingness, and realness. This discrepancy is likely due to its higher scene change rate, which reflects more fragmented sequences and quicker transitions between clips. We encourage the readers to watch the video samples on our demo website.\cref{fn:demo}

\begin{table}
\footnotesize
\centering
\vspace{-3ex}
\caption{Subjective evaluation results of the LLM-generated narrations}
\vspace{1ex}
% inference model: 45 epochs, no diffusion prior
\label{tab:narr_subjective}
\begin{tabular}{lccc}
\toprule
Narration &Organization$\uparrow$ &Informativeness$\uparrow$ &Engagingness$\uparrow$\\ 
\midrule
Naive summarization &3.58 $\pm$ 0.57 &3.72 $\pm$ 0.47 &3.60 $\pm$ 0.56\\
Finely-tuned scripts &\textbf{3.88 $\pm$ 0.44} &\textbf{3.82 $\pm$ 0.54} &\textbf{3.70 $\pm$ 0.46}\\
\bottomrule
\end{tabular}
\end{table}

\begin{table}
\footnotesize
\centering
\vspace{-3ex}
\caption{Effects of the smoothing mechanism and diffusion prior for TeaserGen-LR}
\vspace{1ex}
\label{tab:ablation}
\begin{tabular}{lcccccccc}
\toprule
Decoding &DP &Smoothing &F1 (\%)$\uparrow$ &REP (\%) &SCR (\%) &CLIPScore &VTGHLS\\
\midrule
Greedy &\no &\no &0.53 &81.32 &27.38 &0.59 &0.74\\
Greedy &\no &\yes &1.56 &31.97 &27.18 &0.58 &0.74\\
Greedy &\yes &\yes &1.38 &26.83 &35.47 &0.62 &0.78\\
Beam search &\no &\yes &\textbf{1.88} & 24.16 &41.97 &0.58 &0.74\\
Beam Search &\yes &\yes &\textbf{1.88} & 19.39 &46.56 &0.63 &0.77\\
\bottomrule
\end{tabular}
\end{table}

% \begin{table}
% \scriptsize
% \centering
% \vspace{-3ex}
% \caption{Ablation study on the effects of the smoothing mechanism and the diffusion prior model}
% \vspace{1ex}
% \label{tab:ablation}
% \begin{tabular}{llllcccccccc}
% \toprule
% Model &Query &Decoding &DP &Smoothed &F1 (\%)$\uparrow$ &REP (\%) &SCR (\%) &CLIPScore &VTGHLS\\
% \midrule
% TeaserGen-LR &Narration &Greedy &\no &\no &0.53 &81.32 &27.38 &0.59 &0.74\\
% TeaserGen-LR &Narration &Greedy &\no &\yes &1.56 &31.97 &27.18 &0.58 &0.74\\
% TeaserGen-LR &Narration &Greedy &\yes &\yes &1.38 &26.83 &35.47 &0.62 &0.78\\
% TeaserGen-LR &Narration &Beam search &\no &\yes &\textbf{1.88} & 24.16 &41.97 &0.58 &0.74\\
% TeaserGen-LR &Narration &Beam Search &\yes &\yes &\textbf{1.88} & 19.39 &46.56 &0.63 &0.77\\
% \bottomrule
% \end{tabular}
% \end{table}

\subsection{Ablation Study}
\label{sec:ablation}

\textbf{Experiment on changing the matching score function}\quad
 In this experiment, we compare using CLIPScore \citep{clipl14} versus VTGHLS \citep{lin2023univtg} as the matching metric in \cref{sec:pretraining_based} to examine the effects of changing the matching score function. Unlike CLIPScore, VTGHLS consider both the importance of a frame and the text-visual correspondence. The goal of this experiment is to examine whether incorporating the importance of a frame is beneficial for video-narration matching.
 % should be considered along with text-visual correspondence in teaser generation.
 As shown in \cref{tab:gpt_objective}, we find that the model that uses VTGHLS as the matching score function outperforms that uses CLIPScore instead, resulting in higher F1 score, lower repetitiveness and lower scene change rate, closer to that of the ground truth. This highlights the benefits of using a matching score that takes into account the importance of a frame.
 % This suggests indicates the importance of the frame is an essential factor during frame selection in teaser generation. 
 % with foreground and background factor considered. 

\textbf{Effectiveness of the proposed prompting approach for teaser narration generation}\quad
In this experiment, we examine the teaser generation approach proposed in \cref{sec:narration_generation}.
% Our carefully designed prompts lead to more teaser-like summaries.
We conduct a subjective listening test to compare the teaser narrations generated by a naive summarization prompt and those generated by our proposed prompting approach in terms of organization, informativeness and engagingness (see \cref{sec:narration_evaluation_questions} for the questions we ask in the survey). We report in \cref{tab:narr_subjective}, the mean values from subjective tests, along with the 95\% confidence intervals. The results indicate that our finely-tuned prompts outperform naive summarization methods across all three dimensions. This indicates story-like conversion and ending questions can make narrations scripts closer to human expectations. We provide in \cref{narrationcomp} examples of the 
% we also compare
extractive summary, ground truth narration, LLM with non-finetuned prompts and LLM with finetuned prompts.

% \textbf{Effects of the Smoothing Mechanism}\quad
% \cref{tab:ablation} shows that applying smoothing can significantly decrease the repetitiveness as well as increase the F1 score, demonstrating the effectiveness of the smoothing mechanism proposed in \cref{sec:learning}.

\textbf{Effects of the smoothing mechanism and the diffusion prior model}\quad
% The diffusion prior plays a key role in reducing repetitiveness while boosting the clip ratio and CLIPScore.
In this experiment, we examine the effectiveness of the smoothing mechanism and the diffusion prior on the teaser generation approach proposed in \cref{sec:learning}. \cref{tab:ablation} shows that applying smoothing mechanism can significantly decrease the repetitiveness as well as increase the F1 score, demonstrating the effectiveness of the smoothing mechanism proposed in \cref{sec:learning}. Further,
\cref{tab:ablation} shows that applying diffusion prior leads to lower repetitiveness, higher language-vision correspondence (as indicated by higher CLIP and VTGHLS). The decreased repetitiveness possibly result from the sampling process in diffusion prior, which is a generative model that generate different image embeddings for the same input text embeddings. The increased CLIPScore possibly come from the fact that diffusion prior can bridge the gap between textual embeddings and visual embeddings. Comparing the two results of TeaserGen-LR using beam search decoding, we also find applying diffusion prior can result in a higher scene change rate, partially because diffusion prior would further encourage scene changes.  

In addition to the above ablation study, we provide additional experiment results for ablation study on $\lambda$ and inferring with ground truth narration in \cref{Regularization,GTinference}, respectively. We also provide qualitative examples on our demo page. 

% \paul{show some good and bad examples of model generations? more qualitative analysis}

\section{Discussion and Limitations}
\label{sec:discussion}

% \paul{paragraph is too dense, make into point form. also phrase more positively as future work instead of attacking what we did.}
\textbf{Alignment between objective and subjective evaluation results}\quad
From \cref{tab:gpt_objective,tab:gpt_subjective}, we find that the scene change rate metric aligns well with the coherence and engagingness scores in our subjective evaluation, where a lower scene change rate is preferable. Several survey participants also pointed out that too frequent scene changes negatively affect the viewer experience.
% can be proven with coherence score, which .
% Comparing \cref{tab:gpt_subjective} and \cref{tab:gpt_objective}, we find those with higher CLIPScore has lower coherence score.
Further, we notice that higher CLIPScore does not always lead to
% necessarily mean
a higher alignment score in human evaluation. This is possibly 
% probably due to the fact thatS
because CLIPScore is calculated on the frame-level, while humans perceive narration-video correspondence on the clip-level.
% in an interval-level.
% Therefore, an interval-based narration-video correspondence score are necessary. 
This necessitates a clip-level narration-video correspondence metric for objective evaluation in future work, which is not trivial as it inherently involves scene change detection and clip-level video understanding.

\textbf{Limitations and future work}\quad
% We acknowledge certain limitations in our work.
Finally, we want to point several notable limitations of our work. First, our proposed model leverage several assumptions that might not hold for other media. For example, we assume that a scene change always happens when we move from one sentence to next in the narration. In addition, the our proposed two-stage approach inherently assume that narration plays a more significant role than visual content, which is a reasonable assumption for many documentaries but might not work for more visual-centered media such as vlogs and silent movies. 
Second, the proposed narration-video matching approach cannot accurately match interview scenes commonly seen in documentaries as the pretrained language-vision model cannot associate names with their corresponding faces, which would require a separate video understanding module. Third, teaser narration generation is a creative process. While we have proposed several prompting strategies for generating better teaser narrations, the proposed method still falls short in terms of artistic quality and creativeness compared to scripts created by professionals. Last but not least, we only consider documentary teaser generation in this paper due to the scarcity of public datasets for other media. However, to examine the generalizability of our proposed models, we apply our proposed models to other media in a zero-shot setting, including old movies and educational videos. We provide some qualitative examples of the generated teasers on the demo website.\cref{fn:demo} For future work, we would like to further scale up the dataset and further explore end-to-end approaches for teaser generation. Moreover, we would also like to explore generating music and sound effects to accompany the generated teasers.

\section{Conclusion}

We have presented a new documentary dataset, DocumentaryNet, that consists of 1,269 documentary-teaser pairs with multimodal data streams of video, speech, music, sound effects, narrations and tags. With DocumentaryNet, we have proposed the TeaserGen system for generating teasers for long documentaries. We adopt a narration-centered approach and approach teaser generation by first generating the narration using a pretrained LLM and then selecting accompanying video clips from the main content. We have explored two approaches for narration-video matching: TeaserGen-PT is based on a pretrained contrastive language-vision model and a thresholding mechanism, whereas TeaserGen-LR is a learning-based model that directly models the mapping between the narrations and visuals.
% TeaserGen-PT computes the video-narration matching score and uses a thresholding mechanism to find the accompanying video clips that sums to the desired length; TeaserGen-LR learns the mapping between the narrations and visuals in an embedding space and decodes the generated embeddings by optimizing through beam search.
Through objective and subjective evaluation, we have demonstrated the effectiveness of the proposed system against several baseline models. We hope our work pave a pathway towards long-range multimodal modeling by exploring this new task of documentary teaser generation.

% In this work, we have proposed a new multimodal dataset, \textit{DocumentaryNet}, and a novel task: generating short teasers from long-form documentaries. We propose an pretraining-based approach TeaserGen-PT and a learning-based context-aware approach TeaserGen-LR. 
% We propose three decoding methods. Experiments show that each method has its own strengths and limitations while decoding with beam search has a better overall performance. 
% TeaserGen-PT achieves superior performance in terms of coherence, alignment, engagingness and realness in subjective test. 
% We also explore modules like diffusion prior, smoothing mechanism and regularization to control over repetitiveness and clip fragmentation. Diffusion prior has proven to be effective in reducing repetitiveness by finding more natural cuts within a sentence. In addition, it bridges the gap between textual and image embeddings by first converting textual embeddings to image embeddings with a diffusion model. Smoothing mechanism has also proven to be effective in reducing repetitiveness and increasing coherence in human studies.

\section*{Ethics Statement}\label{ethic}

% Copyright issue
We note that, as a generative model trained on copyrighted material, our proposed system has the potential to generate samples that could lead to copyright infringement. 
% However, considering the potential application in education content preview and accessibility to enhance human experience, we would think teaser generation as a practical and important application.
Moreover, as discussed in \cref{sec:discussion}, the proposed system can make mistakes in matching names and faces, and may pair irrelevant visuals to the generated narrations, resulting in risks of factual inaccuracy.

\section*{Reproducibility Statement}\label{reproduce}

For reproducibility, we will release all the source code, pretrained models and hyperparameters upon acceptance.
% In addition, we will release the metadata as well as the video names and annotations upon acceptance.
Due to copyright concerns, we will release the YouTube URLs for the videos, along with the metadata and annotations. The video files will be made available for research purpose.

\bibliography{iclr2025_conference}
\bibliographystyle{iclr2025_conference}

\appendix

\section{Details of the DocumentaryNet Dataset}\label{datasetdetail}

\textbf{Three-stream Sound Separation}\quad
We extract the audio track of the documentary using the ffmpeg library. We then use a pretrained three-stream audio separation model \citep{audiosep} to separate each audio track into three stems: \textit{dialogue}, \textit{music} and \textit{sound effects}. Due to the length of the audio track of the documentary, we split each documentary into 60-second chunks and then run the separation model chunk by chunk. Finally, we smooth the transitions with a window size equal to 5\% of the audio sampling rate (44,100 Hz in this work). While we do not use the music and sound effect tracks in this work, we include them in the DocumentaryNet dataset as we believe they would be helpful for future research on teaser generation.

\textbf{Silence Detection}\quad
we use a silence detector \citep{silence} to detect periods of silence within each separated audio track. We randomly select 20 samples to select the optimal threshold value for classifying segments as noise or sound.
A binary label is assigned to each segment, with `1' indicating sound and '0' indicating silence. The final thresholds used are -40 dB for the music track, -25 dB for the dialogue track, and -40 dB for the sound effects track.

% \section{GPT Prompts}
% \label{gpt_prompts}

% We assign system\_prompt to be "You are a narrator for this story."

% \begin{itemize}
%     \item \textbf{Prompt to generate summarization}: ``Summarize the paragraph in one sentence''
%     \item \textbf{Prompt to generate story-like narration}: ``Rewrite the paragraph into an engaging story opening in 10 sentences or less, keeping all names and avoiding being replaced by pronouns.''
%     \item \textbf{Prompt to generate ending questions}: ``Given the title and the provided summary, formulate one thought-provoking and concise question that relate directly to the summary.''
% \end{itemize}

% \section{GPT Prompts} \label{sec:gptprompt}

% We use the following prompts to generate teaser narrations using GPT-4o \citep{gpt-4o}, as described in \cref{sec:narration_generation}.

% \begin{itemize}
%     \item \textbf{Story-Like Conversion}: "Rewrite the paragraph into an engaging story opening in 10 sentences or less, keeping all names and avoiding being replaced by pronouns."
%     \item \textbf{Ending Question Generation}: "Given the title and the provided summary, formulate one thought-provoking and concise question that relate directly to the summary"
% \end{itemize}

\section{Comparison of the Generated Narrations}\label{narrationcomp}
We compare original narration, LLM-generated teaser narrations, LLM-generated teaser narrations with finetuned prompts in \cref{tab:narrations,tab:narrations_2}. We leverage T5 (Text-to-Text Transfer Transformer) model to generate extractive summary on transcribed narration. 

\section{Details of the Evaluation Metrics} \label{evaluationdetail}

\textbf{Ground truth matching}\quad
In this work, we only consider those non-fully black video clips that can be extracted from body contents. Therefore, we remove those without match. In order to find those without match, we randomly pick 15 documentaries with teaser and body contents frame pairs. Considering the low FPS, there might be some shift when we extract frames. Therefore, due to frame shift, we allow a higher tolerance in distance. We calculate the lowest 20 \%  $L_{2}$  distances between each teaser frame and each body contents frame. The selected lowest distance threshold is 88.92. To find the matched interval, we first extract CLIP feature with CLIP(L-14). We select top 20 images with the highest cosine similarity to find those close in semantic meanings. Then we calculate pixel-by-pixel L2 distance between teaser frame and those 20 images. We pick the one with the lowest distance and smaller than the selected threshold. In order to remove dark frames, we randomly select frames from the body content of 10 documentaries  and calculate the average brightness of their raw pixels. We define the darkest frames as those in the lowest 5\% of brightness.

\textbf{Repetitiveness}\quad
Since we are not able to find some of the frame in body contents due to frame shift, we calculate the lower bound of repetitiveness by considering all non matched frames as different frames. The lower bound is 0.0786. 

\textbf{Scene Change Rate}\quad
The upper bound is calculated by considering all consecutive non matched frames as a new clip as those consecutive non matched frames might contain more than one clip in practice.
We also randomly pick 10 documentaries teaser and calculate number of video clips per teaser. The average number of clips per teaser is around 27.6 \%. 

\section{Subjective Evaluation Questions for Teaser Generation}
\label{sec:evaluation_questions}

We ask the following questions in our subjective evaluation survey to evaluate our proposed models for teaser generation, as described in \cref{sec:subjective} and reported in \cref{tab:gpt_subjective}.

\begin{itemize}
    \item \textbf{Coherence}: ``To what extent do you feel that the sample maintains coherence and a smooth flow, ensuring that each segment transitions logically and the overall experience feels seamless?''
    \item \textbf{Correspondence}: ``To what extend do you feel that the narration and a video match and work well together, making the overall presentation clear and easy to follow?''
    \item \textbf{Engagingness}: ``To what extent do you feel that the sample captures your interest and keep you engaged throughout?''
    \item \textbf{Realness}: ``How well do you feel this sample meets your expectations as a teaser for a documentary in general?''
\end{itemize}

\section{Subjective Evaluation Questions for Teaser Narration Generation}
\label{sec:narration_evaluation_questions}

We ask the following questions in our subjective evaluation survey to evaluate our proposed approach for generating teaser narration, as described in \cref{sec:narration_generation} and reported in \cref{tab:narr_subjective}.

\begin{itemize}
    \item \textbf{Organization}: To what extent do you feel that the narration present a well-organized and consistent storyline?
    \item \textbf{Informativeness}: To what extent do you feel that the narration introduces the main characters and conflicts clearly and concisely?
    \item \textbf{Engagingness}: To what extend do you feel the narration keeps you engaged and builds curiosity about what happens next?
\end{itemize}

% \begin{table}
% \scriptsize
% \centering
% \caption{Finetune Objective Evaluation Results}
% \label{tab:finetune}
% \begin{tabular}{llllcccccccc}
% \toprule
% Model &Query &Decoding & Finetune &P (\%)$\uparrow$ &R (\%)$\uparrow$ &F1 (\%)$\uparrow$ &REP (\%) &SCR (\%) &ClipS &VTGHLS &ClipSS\\
% \midrule
% \textbf{Interval-based}\\
% TeaserGen-PT-UniVTG &Title &\no &- & 7.06  &1.07 &1.85 &0 &13.16 &0.22 &1.02 &0.56\\
% TeaserGen-PT-UniVTG &Narration &Thresholding & \no &4.45 &0.61 &1.07 &21.38 &22.58 & 0.23 &1.45 &0.58 \\
% TeaserGen-PT-UniVTG & Narration & Thresholding   & \yes& 5.13 & 0.70 & 1.23 & 41.39 &   24.01 & 0.2259 & 1.0533 & 0.5648 \\
% \bottomrule
% \end{tabular}
% \end{table}

% \subsection{Subjective Results}

% \begin{table}
% \scriptsize
% \centering
% \caption{Effect of the regularization term $\lambda$ in \cref{eq:score}}
% % inference model: 45 epochs, no diffusion prior
% \label{tab:lambda}
% \begin{tabular}{lccccccccc}
% \toprule
% Decoding &$\lambda$ &F1 (\%)$\uparrow$ &REP (\%) &SCR (\%) &CLIPScore &VTGHLS\\
% \midrule
% % Beam Search  & 0.1  & 5.6093 & 0.8907 & 1.5372 & 0.2831 & 0.3192 & 0.2359&0.7353 & 0.5896 \\
% % Beam Search  & 0.5  & 5.4159 & 0.8600 & 1.4842 & 0.2825 & 0.3222 & 0.2360 &0.7363 & 0.5899 \\
% Beam search &1   &1.88 &24.16 &41.97 &0.58 &0.74\\
% Beam search &10  &1.91 &24.01 &41.78 &0.58 &0.74\\
% Beam search &100 &1.91 &24.06 &41.74 &0.58 &0.74\\
% % Beam Search(DP)  & 0.5 & 6.0928 & 0.9674  & 1.6698 & 0.2379   & 0.3040  & 0.2306 & 0.2833 & 0.5764 \\
% \bottomrule
% \end{tabular}
% \end{table}

\begin{table}
\scriptsize
\centering
\caption{Objective evaluation results for the experiment on inference with ground truth narrations}
\vspace{1ex}
\label{tab:GT_ablation}
\begin{tabular}{lllcccccc}
\toprule
Backbone &Query &Decoding &Smoothing &F1 (\%)$\uparrow$ &REP (\%) &SCR (\%)  &CLIPScore &VTGHLS\\ 
\midrule
\textbf{Baseline models}\\
Random & Random & - & - & 1.65 & 1.60 & 12.92 & 0.54 & 0.64\\
UniVTG (\citeyear{lin2023univtg}) & Title & Rank & - & 1.67 & 0 & 88.50 & 0.55 & 0.79\\
CLIP-it (\citeyear{clipit}) & Narration & Rank  & - & 1.20 & 0  & 99.24 & 0.54 & 0.44\\
\cmidrule(lr){1-9}
\multicolumn{2}{l}{\textbf{Pretraining-based models}}\\
TeaserGen-PT & Title & Threshold & - &1.46 &0 &17.33 &0.55 & 1.03\\
TeaserGen-PT & Narration & Threshold & - &1.20 &6.12 &24.42 & 0.55 & 1.40\\
TeaserGen-PT-CLIP &Narration & Threshold  & -  &  1.90 &  7.79 &23.85 & 0.56 &0.72\\
% UniVTG  & Narration & Threshold & \yes  & - & -& 2.6112  & 0.9119 & 1.3517 & 0.2007  & 0.2666 & 0.2199 & 0.3450 & 0.5498 \\
\cmidrule(lr){1-9}
\multicolumn{2}{l}{\textbf{Learning-based models}}\\
TeaserGen-LR & Narration& Greedy & \no & 0.88 & 68.87 & 34.19 & 0.57 &0.71\\
TeaserGen-LR &Narration & Greedy  & \yes  & 1.76 & 11.91 & 34.07 & 0.56 & 0.71\\
TeaserGen-LR & Narration & Beam search  & \yes  & 2.20 & 10.16 & 45.84 & 0.56 &0.72\\

\cmidrule(lr){1-9}
Ground truth &- &- &- &100 &$>$7.86 &27.6 &0.58 &0.64\\
\bottomrule
\end{tabular}
% \\[1ex]
% \tabfn \tbaa{did we want to say something here??} \quad 
% \tabfn[2] We manually inspect 10 videos to estimate the scene change rate for the ground truth teasers. 
\end{table}

\begin{table}
\footnotesize
\centering
\caption{Subjective evaluation results for the experiment on inference with ground truth narrations}
\vspace{1ex}
\label{tab:gt_subjective}
\begin{tabular}{lllc@{~~~}c@{~~~}c@{~~~}c}
\toprule
Model &Query &Decoding &Coherence$\uparrow$ &Correspondence$\uparrow$ &Engagingness$\uparrow$ &Realness$\uparrow$\\
\midrule
UniVTG {\scriptsize(\citeyear{lin2023univtg})} &Title &Rank &2.64 $\pm$ 0.53 &2.72 $\pm$ 0.60 &2.76 $\pm$ 0.53 &2.64 $\pm$ 0.57\\
% CLIP-it\citep{clipit} & Narration & Rank &\no &\no &  \\
% CLIP & Original & Narration & \no & 2.6 & 0.1 & 1.5 & \textbf{TODO} & 10 &  4.3 \\
TeaserGen-PT &Title &Threshold &\textbf{3.45 $\pm$ 0.47 }  &\textbf{2.96 $\pm$ 0.66}  & \textbf{2.91  $\pm$ 0.58} & \textbf{2.97$\pm$ 0.63} \\
TeaserGen-LR &Narration &NN &2.93 $\pm$ 0.58 &2.69 $\pm$ 0.62 &2.74 $\pm$ 0.62 &2.61$\pm$ 0.67\\
% Transformer &Narration & Beam Search  & - & \no  & \\
% Transformer &Narration & Beam Search  & - & \yes &\\
% \cmidrule(lr){1-9}
% Ground Truth & - &  - & - & -& \\
\bottomrule
\end{tabular}
\end{table}

\section{Experiment on Inference with Ground Truth Narrations}\label{GTinference}

Although we cannot have ground truth narration when making teasers in reality, we conduct an ablation study to compare model behavior on real narrations and machine generated narrations. We did not apply the diffusion prior model for TeaserGen-LR in this experiment. We report objective evaluation results in \cref{tab:GT_ablation} and subjective evaluation results in  \cref{tab:gt_subjective}. As shown in \cref{tab:gt_subjective}, TeaserGen-PT achieves the highest score in terms of coherence, correspondence and engagingness in human evaluation. 

\begin{table}
\footnotesize
\centering
\caption{Objective evaluation results for the finetuning experiment with TeaserGen-PT}\vspace{1ex}
\label{tab:finetune}
\begin{tabular}{lllccccccc}
\toprule
Query &Decoding &Finetune &F1 (\%)$\uparrow$ &REP (\%) &SCR (\%) &CLIPScore &VTGHLS\\
\midrule
% \textbf{Interval-based}\\
Title & Thresholding & \no  &1.85 &0 &13.16 &0.56 &1.02\\
Narration &Thresholding & \no  &1.07 &21.38 &22.58 &0.58 &1.45\\
Narration & Thresholding & \yes & 1.23 & 41.39 &24.01 &0.56 &1.05\\
\bottomrule
\end{tabular}
\end{table}

\section{Experiment on Finetuning the Pretrained Video Temporal Grounding Model}\label{finetune}

% \paragraph{Data Preparation}

We prepare highlight detection dataset by constructing query and highlight frame pairs with timestamped narration. We have around 10,333 samples to finetune the pretrained UniVTG. However, \cref{tab:finetune} shows the finetuned model result in a worse performance due to the relatively small size of the finetuning dataset comparing with the dataset in pretraining.

\section{Effects of the Regularization Term $\lambda$} \label{Regularization}

We show in \cref{tab:lambda} the results when using different values of $\lambda$ in \cref{eq:score}. As shown in \cref{tab:lambda}, we find that a higher $\lambda$ can reduce the repetitiveness and lead to a higher F1 score but the difference is not significant. 

\begin{table}
\footnotesize
\centering
\caption{Effect of the regularization term $\lambda$ in \cref{eq:score}}
\vspace{1ex}
% inference model: 45 epochs, no diffusion prior
\label{tab:lambda}
\begin{tabular}{lccccccccc}
\toprule
Decoding &$\lambda$ &F1 (\%)$\uparrow$ &REP &SCR &CLIPScore &VTGHLS\\
\midrule
% Beam Search  & 0.1  & 5.6093 & 0.8907 & 1.5372 & 0.2831 & 0.3192 & 0.2359&0.7353 & 0.5896 \\
% Beam Search  & 0.5  & 5.4159 & 0.8600 & 1.4842 & 0.2825 & 0.3222 & 0.2360 &0.7363 & 0.5899 \\
Beam search &1   &1.88 &24.16 &41.97 &0.58 &0.74\\
Beam search &10  &1.91 &24.01 &41.78 &0.58 &0.74\\
Beam search &100 &1.91 &24.06 &41.73 &0.58 &0.74\\
% Beam Search(DP)  & 0.5 & 6.0928 & 0.9674  & 1.6698 & 0.2379   & 0.3040  & 0.2306 & 0.2833 & 0.5764 \\
\bottomrule
\end{tabular}
\end{table}

\begin{table}
    \scriptsize
    \centering
    \caption{Comparison of LLM-generated teaser narrations---Example I}
    \vspace{1ex}
    \label{tab:narrations}
    \begin{tabularx}{\linewidth}{p{.4in}X}
        \toprule
        Model &Narration\\
        \midrule
        Extractive & archaeologists are searching for clues to reveal more about the pharaoh queen. hetshepsood's temple is the key to the secrets of her dynasty, says dr zbigniew szybranski, a doctor of ranski team. the temple was buried and badly damaged and little was known about its owner. a ram-headed sphinx has been excavated at one of the largest archaeological sites in ancient Egypt. it's thought that Hedge Sheppset started the greatest display of Sphinces known to ancient Egyptians. the necropolis of around 100 ancient tombs is 120 miles from Luxor in Aswan. cnn's richard quest is excavating what could be one of the sphinxes of Hatshepsut. quest finds a small cryo-like body in the quarry at gopel, in southern egypt, near the tomb of king henry ibn al-qaeda - he was buried there in 1586 b.c. doctors of ransky is investigating the paintings she left on the walls of head chef's at temple. the imagery holds clues to her life and reveals a family power struggle with her nephew, who is also her stepson, Tutmost III. despite being female, Hadshepsut is often depicted as a man. but hidden in the hieroglyphic text, Colleen finds evidence of her real gender. at the Necropolis in Aswan, a team finds an ancient face mask made of cartonage. it was skillfully how they made the eyes of an adult mummy buried outside the tomb. the mask was meant to help the afterlife from gold, this one is more beautiful than to come once again. american archaeologist finds dramatic evidence of one individual's life and death. the vagina is the whole here, still very distended, so we know within 24 hours of giving birth, that she died. but the female pharaoh had chefsuit wanted to be more than just equal. doctors of Ransky are working to restore temple of Hachepsut. ancient earthquakes and landslides have damaged the temple's upper terrace, which is 11 feet higher than the valley of the kings. the bones and delicate cartonage have been moved into a lab in aswan, savan. hundreds of sphinxes still remain at carnax temple site. a ten-foot high abandoned statue is the largest they've seen at the quarry, john and maria Ward say. the couple are looking for black inclusions within the stone itself. there are no records whatsoever of any unfinished statue that is intact like this. sphinxes from sosilla were shipped 100 miles down the nile to carnac temple. the evidence suggests they came from the city, not so far from where they are now - john w. mcdonald jr.\\
        \cmidrule(lr){1-2}
        LLM-generated &
        Archaeologists, led by Dr. Zbigniew Szybranski, are excavating and restoring the unique temple of the revolutionary female pharaoh Hetshepsood in Luxor, Egypt, to uncover more about her enigmatic reign and contributions, while other teams investigate her quarries to understand her extensive building projects
        John and Maria are excavating an abandoned statue at one of Egypt's largest archaeological sites, while Martina Bartanova's team in Aswan discovers ancient human remains that could halt their exploration of an unopened tomb.
        Martina and her team are excavating a site where they discover a child's remains and a miniature sphinx, while dealing with the challenges of preserving delicate artifacts and documenting their findings for the Egyptian government. 
        The paragraph describes various archaeological efforts in Egypt, including the discovery of a miniature model likely carved for practice, the investigation of Queen Hatshepsut's rise to power and her conflict with her stepson, the challenges faced by a team uncovering artifacts near her temple, and the work of Yale professors using digital technology to study ancient inscriptions.
        The couple explores Karnak Temple to investigate inscriptions revealing Hadshepsut's portrayal as a male pharaoh, while archaeologists at Dra'abu El Naga study the roles of women in ancient Egyptian society through the excavation of a tomb filled with fragmented human remains.
        The team is meticulously organizing human remains, including mummified organs and detached body parts, while making significant discoveries such as a young woman who likely died in childbirth and an ancient face mask, all under the pressure of impending strong winds.
        American archaeologist Susanne Onstein is uncovering the roles and mortal dangers faced by women in ancient society, while also exploring the monumental building campaigns of female pharaoh Hatshepsut, who used colossal obelisks as both religious devotion and propaganda to immortalize her reign.
        The excavation team has unearthed ancient pottery fragments and evidence of temples built by Hatshepsut and her stepson Tutmose III, revealing a historical power play and ongoing restoration efforts, while another team in Aswan uncovers Middle Kingdom burials, all amidst the personal reflections and challenges faced by the archaeologists.
        John and Maria unearthed a nearly 10-foot-tall abandoned sphinx statue at a quarry and are now investigating at the Carnax temple to find a matching sphinx, focusing on identifying black inclusions in the sandstone to determine its origin.
        John and Maria's discovery of sphinxes from Sosilla, transported to Carnac Temple, alongside Dr. Ransky's excavation and restoration efforts, has unearthed significant artifacts and revived the legacy of the remarkable Pharaoh Queen Hatshepsut.\\
        \cmidrule(lr){1-2}
        LLM-generated (tuned) &Under the scorching Egyptian sun, Dr. Zbigniew Szybranski led a team of archaeologists in Luxor, meticulously excavating the unique temple of the revolutionary female pharaoh Hetshepsood. The air buzzed with excitement as John and Maria unearthed a nearly 10-foot-tall abandoned sphinx statue at one of Egypt's largest archaeological sites. Meanwhile, in Aswan, Martina Bartanova's team stumbled upon ancient human remains that could halt their exploration of an unopened tomb. The discovery of a child's remains and a miniature sphinx added layers of mystery to their quest. At Karnak Temple, John and Maria delved into inscriptions revealing Hetshepsood's portrayal as a male pharaoh, while Yale professors used digital technology to study ancient texts. In Dra'abu El Naga, another team meticulously organized fragmented human remains, uncovering stories of women in ancient Egyptian society. American archaeologist Susanne Onstein explored the monumental building campaigns of Hetshepsood, whose colossal obelisks stood as testaments to her reign. Amidst pottery fragments and evidence of temples built by Hetshepsood and her stepson Tutmose III, the archaeologists faced personal reflections and challenges. The winds of history whispered through the sands, as each discovery brought them closer to unraveling the enigmatic legacy of Pharaoh Queen Hetshepsood. How do the discoveries and restoration efforts at Hatshepsut's temple and other archaeological sites in Egypt contribute to our understanding of her reign and the broader role of women in ancient Egyptian society?\\
        \cmidrule(lr){1-2}
        Ground truth &
        Egypt, the richest source of archaeological treasures on the planet, beneath this desert
         landscape, why the secrets of this ancient civilization?
         Wow!  You can see why a Pharaoh's chosen place, now, for a full season of excavations, our  cameras have unprecedented access, follow teams on the front line of archaeology.  I'm writing so fast because I'm subsided!
        It's an entrance, we can see an entrance. Revealing varied secrets.
        I have just been told that they have found something
         Making discoveries, they could rewrite ancient history.
        This time, new secrets about one of Egypt's greatest rulers, the Pharaoh Queen,  Hatshepsut, doctors of Ranski discovers very treasures that her magnificent temple  had reached was to be remembered for millions of years.  The Darnels uncovered how she formed a mysterious double identity to seize power.  For my beloved daughter, not son, and John and Maria, honor a rare and intriguing statue.\\
        \bottomrule
    \end{tabularx}
\end{table}

\begin{table}
    \scriptsize
    \centering
    \caption{Comparison of LLM-generated teaser narrations---Example II}
    \vspace{1ex}
    \label{tab:narrations_2}
    \begin{tabularx}{\linewidth}{p{.4in}X}
        \toprule
        Model &Narration\\
        \midrule
        Extractive & the mountains here were a battleground when US and NATO troops were still in the country. outside bombs and airstrikes have left their mark. the damage lines the road. in her large garden, Dr. Rochanak Wardak takes us down memory lane. gynecologist dr. rochanak's patients are still suffering the deprivation of that war. her shoulders aren't neck hurt. it makes no difference if I eat or not. with the Taliban, things are good now. it's peaceful. there was a lot of war and fighting. every pregnant woman wants to know the sex of her baby. yes, boy is power of the family. the drive through Kabul will reduce more drug users and street children than we've seen before. more than half of the population is hungry; food prices are soaring. women sit in front of bakeries, begging for a loaf of bread. but it's not enough to fill our stomachs. frida ghitis: the hunger crisis is on the government of the Taliban. she says they want a gender segregated society, but girls dare to dream of being able to do the same. aisha and her sister are trying to find their place in this Afghanistan. she says she has lost her spirit since the Taliban returned to power. aisha khan: if i were president, I would form an all-female government. she says women would be allowed to choose any job they wanted, study and live freely. her country is tired. tired of betrayal. you are songless and voiceless. cnn's aisha nasri meets with teen who wants to join the Taliban to serve the Islamic Emirate. she says many Afghans fear Taliban rule, but he and his friends brush aside such concerns, she adds. they say talk of explosions, targeted killings and arbitrary arrests is nothing more than Western propaganda. the traditional pottery of Istarlif was so foreign as darling when u.s. troops were still in the country. the pressure on visitors has affected the potters of the mountain village, which is now closed to tourists and a monopoly by the taliban. if no one buys their wares, they may not be able to do what they have always done. "we're afraid of new wars and more hardship," says Abdul manan. he says the traditional life has survived much, but now feels under threat again.\\
        \cmidrule(lr){1-2}
        LLM-generated &
        We traveled west from Kabul to the war-torn province of Maidan-Wardak to meet Dr. Rochanak Wardak, a former politician and Taliban supporter, who reminisces about her efforts to build a new Afghanistan governed by parliament.
        War devastated the country, halting its development and education, leading Dr. Rochanak, a gynecologist who now runs a small home clinic, to witness her patients still suffering from the war's deprivation.
        With the cessation of fighting, travel has become safer, and while the region is now peaceful under the Taliban, there remains a cultural emphasis on the importance of having boys for family power and security, with some boys she helped deliver now becoming fighters. 
        Dr. Rosha Nakh, who works for the Taliban overseeing female and maternal health, hopes for a change in their stance on women's activities, but faces the harsh reality of a male-dominated regime and widespread poverty in Kabul.
        The Taliban blame the hunger crisis on international sanctions and aid cuts, asserting their right to enforce their vision of Islamic rule and demanding autonomy, while their policies focus more on controlling women's lives than addressing economic issues.
        The Taliban's ban on secondary education for girls has left young women like Aisha, who once aspired to be doctors and pursued careers, feeling lost and hopeless as they struggle to find their place in the new Afghanistan.
        Aisha feels like a prisoner and blames the West for abandoning Afghanistan to the Taliban, which has led her to withdraw from public life and dream of an all-female government to ensure women's freedom and opportunities.
        The men would stay at home to experience the restrictive life of Afghan women, while Isatullah, a former farmer turned Taliban member, dismisses fears of Taliban rule as Western propaganda and expresses his support for the Sharia-based Islamic Emirate.
        The Taliban's restrictions on visitors, especially women, have severely impacted the traditional pottery business in the mountain village of Istarlif, leaving craftsmen like Abdul Manan and his son Ramish struggling to sustain their family craft.
        Abdul Manan, a traditional artist, expresses fear for his livelihood and safety amidst ongoing conflicts and deserted marketplaces, highlighting the persistent threat to their peaceful existence despite not taking sides in wars.\\
        \cmidrule(lr){1-2}
        LLM-generated (tuned) &
        The journey west from Kabul to the war-torn province of Maidan-Wardak was fraught with memories of conflict and hope. Dr. Rochanak Wardak, a former politician and Taliban supporter, welcomed us into her small home clinic. Once a gynecologist with dreams of a parliamentary Afghanistan, Dr. Rochanak now witnesses the lingering scars of war in her patients' eyes. The cessation of fighting has brought a fragile peace, but the cultural emphasis on having boys for family power remains. Some of the boys Dr. Rochanak helped deliver are now fighters, a stark reminder of the cycle of conflict. Dr. Rosha Nakh, working for the Taliban in female and maternal health, dreams of a shift in the regime's stance on women's activities. However, the male-dominated reality and widespread poverty in Kabul cast a long shadow over her hopes. The Taliban blame international sanctions for the hunger crisis, focusing more on controlling women's lives than addressing economic issues. Young women like Aisha, once aspiring doctors, now feel lost and hopeless, their dreams crushed by the Taliban's ban on secondary education for girls. In the mountain village of Istarlif, traditional potter Abdul Manan and his son Ramish struggle to sustain their family craft amidst severe restrictions on visitors. Abdul Manan's fear for his livelihood and safety underscores the persistent threat to their peaceful existence, a poignant reminder of the ongoing conflicts that continue to shape Afghanistan's future. How does Dr. Rochanak Wardak reconcile her support for the Taliban with her aspirations for a progressive Afghanistan, especially in light of the Taliban's restrictive policies on women's education and activities?\\
        \cmidrule(lr){1-2}
        Ground truth &
        Taliban fighters rule the streets, armed with weapons once used by their enemies.
         We see them all over Kabul.
        The Taliban's return to power has changed millions of lives, especially the lives of  Afghan women.  I feel like a prisoner, nothing is in my own hands anymore, before I could do whatever I wanted.
         Of course, I'm very much happy.  You know, the head is no war.
        We are living like other people in the world.
        The Taliban are back and so is their ideology, but what's it like living in today's  Afghanistan?\\
        \bottomrule
    \end{tabularx}
\end{table}

\end{document}